\documentclass{article}

\usepackage{microtype}
\usepackage{graphicx}
\usepackage{booktabs} 

\usepackage{hyperref}

\usepackage[accepted]{icml2024}

\usepackage{amsmath}
\usepackage{amssymb}
\usepackage{mathtools}
\usepackage{amsthm}

\usepackage[capitalize,noabbrev]{cleveref}

\theoremstyle{plain}

\theoremstyle{definition}

\theoremstyle{remark}

\usepackage[textsize=tiny]{todonotes}

\usepackage{multirow}
\usepackage{adjustbox}
\usepackage{graphicx}

\usepackage{amsmath}
\usepackage{amsthm}
\usepackage{xspace}
\usepackage{pifont}
\usepackage{wrapfig}
\usepackage{natbib}

\usepackage{colortbl} 

\usepackage{enumerate}
\usepackage{framed}
\usepackage{pifont}
\newcommand{\cmark}{\ding{51}}
\newcommand{\xmark}{\ding{55}}
\usepackage{relsize}
\usepackage{enumitem}
\usepackage{tcolorbox}
\usepackage{listings}
\usepackage{wrapfig}

\newcommand{\llama}{\textsc{Llama2}\xspace}
\newcommand{\model}{\textsc{ChunkLlama2}\xspace}

\icmltitlerunning{Training-Free Context Scaling of Large Language Models}

\begin{document}

\twocolumn[
\icmltitle{Training-Free Long-Context Scaling of Large Language Models}



\icmlsetsymbol{equal}{*}

\begin{icmlauthorlist}
\icmlauthor{Chenxin An}{equal,hku,ali}
\icmlauthor{Fei Huang}{ali}
\icmlauthor{Jun Zhang}{}
\icmlauthor{Shansan Gong}{hku}
\icmlauthor{Xipeng Qiu}{fdu}
\icmlauthor{Chang Zhou}{ali}
\icmlauthor{Lingpeng Kong}{hku}

\end{icmlauthorlist}

\icmlaffiliation{hku}{The University of Hong Kong}
\icmlaffiliation{ali}{Alibaba Group}
\icmlaffiliation{fdu}{Fudan University}

\icmlcorrespondingauthor{Chenxin An}{cxan23@connect.hku.hk}

\icmlkeywords{Machine Learning, ICML}

\vskip 0.3in
]



\printAffiliationsAndNotice{\icmlEqualContribution} 

\begin{abstract}
The ability of Large Language Models (LLMs) to process and generate coherent text is markedly weakened when the number of input tokens exceeds their pretraining length. Given the expensive overhead of finetuning large-scale models with longer sequences, we propose Dual Chunk Attention (DCA), which enables \llama 70B to support context windows of more than 100k tokens without continual training.  By decomposing the attention computation for long sequences into chunk-based modules, DCA manages to effectively capture the relative positional information of tokens within the same chunk (Intra-Chunk) and across distinct chunks (Inter-Chunk), as well as integrates seamlessly with Flash Attention.
In addition to its impressive extrapolation capability, DCA achieves performance on practical long-context tasks that is comparable to or even better than that of finetuned models. When compared with proprietary models, our training-free 70B model attains 94\% of the performance of \texttt{gpt-3.5-16k}, indicating it is a viable open-source alternative.
All code and data used in this work are released at \url{https://github.com/HKUNLP/ChunkLlama}.

\end{abstract}
\section{Introduction}
The ability to comprehend and process long-context information is essential for large language models (LLMs)~\cite{gpt-4, llama, llama2,bai2023qwen, claude} to cater to a wide range of applications effectively. These include analyzing and responding to inquiries within sizable PDFs, retaining extended dialogue history, and empowering interactive chatbots~\cite{wei2023leveraging, Lee_2023,rula2023procedural, saadfalcon2023pdftriage, lv2024longwanjuan}.

Recent advances have shown that the long-context ability can be improved by further training a short-context model on long text sequences~\cite{ruoss2023randomized, codellama}. The impressive performance of Llama2 Long~\cite{llama2long}, which is trained from a mix of long text data and the original Llama2~\cite{llama2} pre-training corpus, stands as a testament to this approach. 
Nevertheless, due to the limited accessibility of these training corpora and the prohibitive cost of long-context finetuning, current open-source models often fall short in performance when compared to the proprietary counterparts, and are generally available in smaller sizes (e.g., 7B/13B).  

Given these constraints, approaches that do not require additional training for context scaling in LLMs become particularly attractive. Recent training-free methods, including LM-infinite~\cite{han2023lminfinite} and StreamingLLM~\cite{xiao2023efficient}, have shown that LLMs trained on a limited context window can efficiently process text of infinite length~\cite{Zhang2023LinearAV, Zhang2024SoaringF4, xiao2024infllm, Qin2024LightningAA}. Assuming that LLMs are unable to generalize to texts longer than the training length, these models handle extended sequences by selectively retaining essential local information. Such paradigms effectively maintain a low Perplexity (PPL), yet they lose long-range dependencies.
To retain the global information, another perspective is to effectively extrapolate to sequence lengths that surpass those encountered during their training~\cite{xpos,kazemnejad2023impact,liu2023scaling, chi2023dissecting}. 
Popular techniques for Llama-based models, including Position Interpolation (PI)~\cite{pi} and NTK-Aware RoPE (NTK)~\cite{fixedNTK, dynamicNTK}, are adaptations of Rotary Positional Encodings (RoPE)~\cite{rope}. These scaled positional encodings necessitate fewer finetuning steps compared to the original RoPE, and their training costs can be further reduced via methods such as YaRN~\cite{yarn} and CLEX~\cite{chen2023clex}.
However, in a training-free setting, we find that these approaches usually lead to a notable increase in PPL especially in input lengths that are more than twice the training length (\S\ref{sec:exp}, Table~\ref{tab:ppl_pg19}).

In this paper, we introduce Dual Chunk Attention (DCA), a new training-free framework to extrapolate the context window of LLMs. We avoid linearly downscaling the position indices or increasing the base frequency in RoPE~\cite{rope}. Instead, we opt to reuse the original position indices with their embeddings from the pretrained model, yet to redesign the construction of the relative position matrix in such a way that it can accurately reflect the relative position of two tokens as faithfully as possible. Inspired by efficient chunk-based attention patterns~\cite{blocksparse, song2023zebra, ratner2023parallel, he2024stones}, DCA segments self-attention computations for a long sequence into small chunks, each chunk being smaller than the size of the pretraining window.  DCA consists of three components: (1) intra-chunk attention, tailored for processing tokens within the same chunk; (2) inter-chunk attention, for processing tokens between distinct chunks; and (3) successive chunk attention, for processing tokens in successive, distinct chunks. These respective treatments help the model effectively capture both long-range and short-range dependencies in a sequence.  In addition to that, the chunk-based attention calculation can be seamlessly integrated with Flash Attention 2~\cite{flashattention, dao2023flashattention2}, a key element for long-context scaling in the open-source community.\footnote{Without Flash Attention, the maximum input tokens for Llama2 7B/13B is about 16k, and for Llama2 70B, it is 5k when tested on two A100 80G GPUs in our experiments}

We present a comprehensive evaluation of our models on a diverse range of tasks that include language modeling, passkey retrieval, and real-world long-context applications that span question answering~\cite{quality,narrativeqa,qasper,leval} and summarization~\cite{qmsum}. Unlike previous work that is usually limited to verification on 7B/13B models, the significant training efficiency of our method makes it possible to validate on 70B models, ensuring robust conclusions. To verify the model's long-context ability independent of potential data exposure during pretraining, we used this paper itself as the input and crafted a series of questions for the models.\footnote{ We invite interested readers to examine the results in Tables~\ref{tab:case1},\ref{tab:case2}}  Our empirical results reveal the following insights:

\begin{enumerate}[leftmargin=0.3cm, itemindent=0.0cm, itemsep=0.0cm, topsep=0.0cm]
\item \textbf{Extrapolation}.
On language modeling, DCA marks a significant advance for training-free approaches. It first shows that  LLMs with a 4k context window can be expanded to more than 32k without training, maintaining a negligible increase in PPL, whereas previous methods typically falter at context lengths beyond 8k. Furthermore, we demonstrate that Llama2 70B, when integrated with DCA, showcases an exceptional extrapolation capability to handle context sizes exceeding 100k tokens. 

\item \textbf{Orthogonality}.
DCA is orthogonal to existing popular scaled positional encodings such as PI~\cite{pi} and NTK~\cite{fixedNTK,dynamicNTK}. We empirically show that existing long-context LLMs, which have already supported a 32k context window, can further extrapolate to a 192k context length while maintaining high passkey retrieval accuracy and low perplexity.

\item \textbf{Long-Context Understanding}. We evaluate DCA on a suite of long-context understanding benchmarks in both zero-shot and few-shot settings. The results suggest that our training-free models achieve performance comparable to, or even surpassing, that of existing state-of-the-art models built through costly continual training. 
\end{enumerate}
\section{Background}
\subsection{Positional Encoding}
The original positional embedding from the Transformer model~\cite{transformers} maps absolute position indices to a $d$-dimensional feature space, and incorporates this into the input layer. The input $\mathbf{x}$, associated with the position index $i$, is expressed as:
$\mathbf{x}_{i} = \mathbf{x} + f(i)$, where $f: \mathbb{N} \rightarrow \mathbb{R}^d$ is the (positional) embedding function.

\begin{figure}
\centering
    \includegraphics[width=0.27\textwidth]{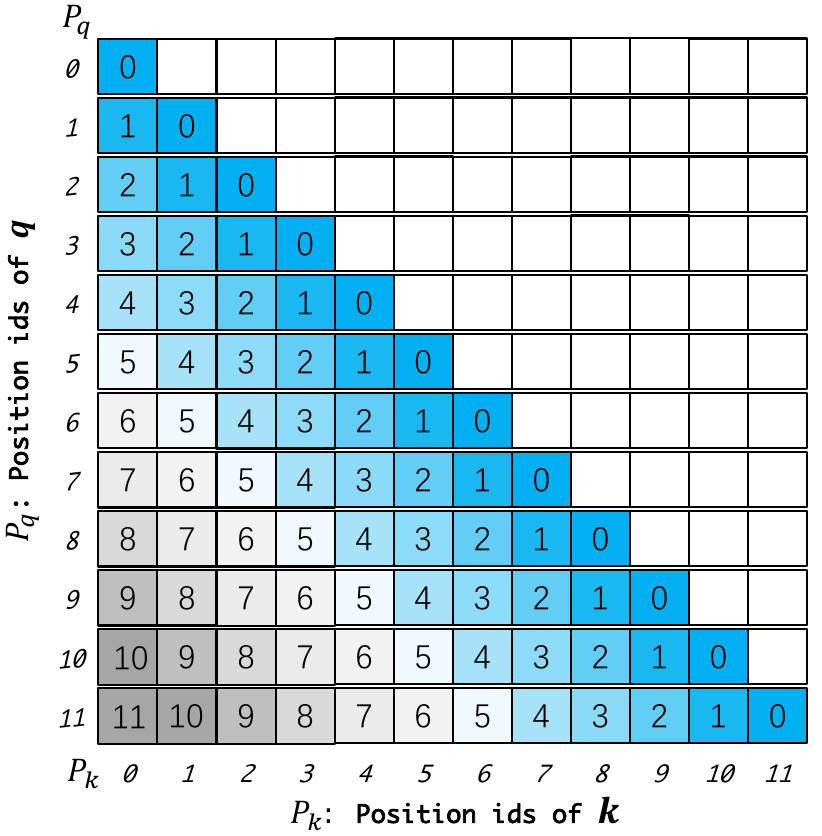}
    \vspace{-1.5em}
    \caption{
    Visualization of the Relative Position Matrix $M$ utilizing standard RoPE. The pretraining context window is 6 and the input sequence length is 12. The x-axis $P_\mathbf{k}$ indicates the position indices of keys, while the y-axis $P_\mathbf{q}$ corresponds to the position indices of queries. Each matrix entry $M[i][j]$ represents the relative positional offset $P_\mathbf{q}[i] - P_\mathbf{k}[j]$.
    }
    \label{fig:rope}
\end{figure}

One of the most prevalent positional encoding methods for LLMs is the Rotary Positional Encoding (RoPE)~\cite{rope}. RoPE eschews the conventional approach of infusing positional information into the input layer. Instead, it directly incorporates this information into the attention layer. For a sequence of $l$ tokens, we denote the position indices for keys and queries\footnote{Queries and keys are usually derived by projecting the input $\mathbf{x}$ through a learnable linear layer.} as follows:
\begin{equation}
    P_\mathbf{k} =P_\mathbf{q} = [0, 1, \ldots, l-1].
\end{equation}
We abuse the notation $f$ for the embedding function of RoPE, which accepts a query vector $\mathbf{q}$ or a key vector $\mathbf{k}$, and the respective position index as arguments. For example, we have $ \mathbf{q}_i = f(\mathbf{q}, P_\mathbf{q}[i])$ and 
$\mathbf{k}_j = f(\mathbf{k}, P_\mathbf{k}[j])$, 
where $[i]$ denotes the $i$-th element of the list. In the most straight-forward case, we have $P[i] = i$. The function $f$\footnote{A typical implementation of $f$ can be found in \href{https://github.com/huggingface/transformers/blob/main/src/transformers/models/llama/modeling_llama.py}{modeling\_llama.py Line 211} \texttt{apply\_rotary\_pos\_emb()}}outputs a modified query or key vector that encapsulates the position index, ensuring that the inner product between the $i$-th query and the $j$-th key (for $i\geq j$) captures the relative positional information $P_\mathbf{q}[i] - P_\mathbf{k}[j]$. 
Although RoPE takes absolute position indices as input, the result of the inner product of $\mathbf{q},\mathbf{k}$ only contains relative position information (i.e., we have $\mathbf{q}_i^\top\mathbf{k}_j = \mathbf{q}_m^\top\mathbf{k}_n$ whenever $m - n = i - j$). The relative position matrix $M$ introduced by RoPE during self-attention can be described as a Toeplitz matrix, as shown in Figure~\ref{fig:rope}. Each element $M[i][j] = P_\mathbf{q}[i] - P_\mathbf{k}[j]$ signifies the relative position between $\mathbf{q}_i$ (the $i$-th query) and $\mathbf{k}_j$ (the $j$-th key). 

\subsection{Extrapolation of RoPE}
\label{ssec:extrapolation}
Recent work~\cite{pi,chowdhury2023monotonic,chen2023clex} has demonstrated that LLMs with the original RoPE lack robust length extrapolation capabilities, typically resulting in performance degradation when tested on input sequences longer than those seen during pretraining~\cite{li2023functional, zhu2023pose}. Recent studies~\cite{pi, rerope2023, selfextend} mainly attribute this limitation to the presence of unseen relative positions in pretraining phase and propose to redesign the relative position matrix.  As illustrated in the example in Figure~\ref{fig:rope}, the model is trained on sequences of 6 tokens, while inference is carried out on a sequence of 12 tokens. This discrepancy can lead to a high PPL because relative positions beyond 6 were never trained. Previous approaches, such as PI and NTK, aim to mitigate this issue by reducing the magnitude of $M[i][j]$ to ensure it falls within the scope of the observed context length during training. For instance, applying PI in this example would adjust the position indices by scaling: $P_\mathbf{q}[i] \Rightarrow P_\mathbf{q}[i] / 2$ and $P_\mathbf{k}[j] \Rightarrow P_\mathbf{k}[j] / 2$. Consequently, the relative position matrix is also scaled: $M[i][j] = M[i][j] / 2$. Here, a scaling factor $2=\frac{12}{6}$ is employed to scale down the relative positions, leading to inferior resolution of the position information and weak extrapolation ability.

\section{Method}
\begin{figure*}
\centering
    \includegraphics[width=0.8\textwidth]{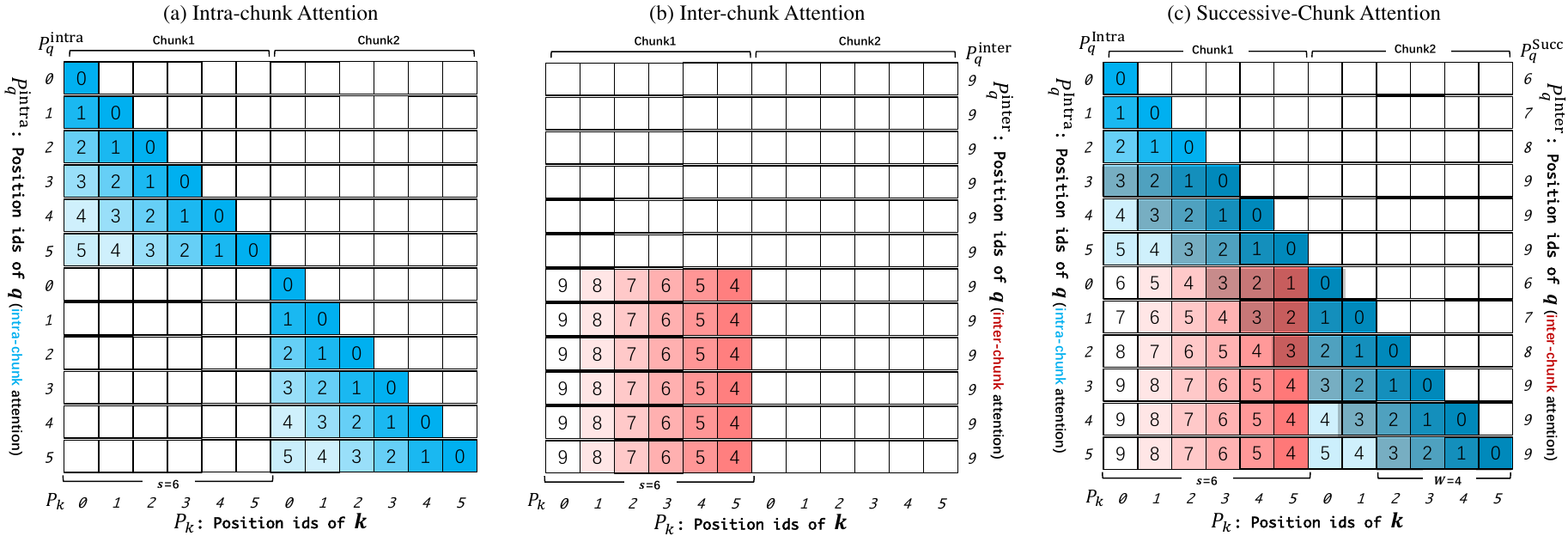}
    \vspace{-0.9em}
\caption{Visualization of the Relative Position Matrix $M$ employing Dual Chunk Attention (DCA), with chunk size $s=6$, pretraining window size $c=10$, and local window size $w=4$ noted by the shadow in (c). The sequence is segmented into chunks to ensure that relative positions do not exceed 9. The matrix element $M[i][j]$ represents the relative position between the $i$-th query vector $\mathbf{q}$ and the $j$-th key vector $\mathbf{k}$. Unlike the original position indices for $\mathbf{q}, \mathbf{k}$ in RoPE, DCA utilizes distinct position index sets $P_\mathbf{k}, P^{\text{Intra}}_\mathbf{q}$ (defined in Eq.~\ref{eq:intra-k}), $P^{\text{Inter}}_\mathbf{q}$ (defined in Eq.~\ref{eq:inter-q}), $P^{\text{Succ}}_\mathbf{q}$  (defined in Eq.~\ref{eq:succ}) to compute the relative distances within different sections of $M$.}
  \label{fig:model}
\end{figure*}
\label{sec:method}
In this section, we describe our new training-free framework Dual Chunk Attention in detail. 
A running example of dual chunk attention is shown in figure~\ref{fig:model}. 
Our method starts from the intra-chunk attention (Figure~\ref{fig:model} (a)) which is a chunk-based efficient attention pattern~\cite{blocksparse, song2023zebra}. The position embedding of each chunk ranges from 0 to chunk size where the chunk size is set to be smaller than pretraining length.
The intra-chunk attention pattern practically means directly truncating the input from left to the chunk size discarding information from previous chunks. Such truncation usually brings low perplexity~\cite{xiao2023efficient} but loses long-range information. To address this limitation, we implement inter-chunk attention (Figure~\ref{fig:model} (b)) that enables attention calculations between different chunks, albeit with less precision for distant token positions. Finally, we introduce successive-chunk attention, a variant of inter-chunk attention depicted in Figure~\ref{fig:model} (c), which is specifically applied when two chunks are adjacent in order to preserve locality.  An ablation study to show how these attention mechanisms influence  PPL and passkey retrieval accuracy can be found in Figure~\ref{fig:ablation}.

\subsection{Intra-Chunk Attention}
Intra-Chunk Attention is employed to calculate the inner product of queries and keys within the same chunk. For a long sequence of length $l$, we partition the sequence into $n = \frac{l}{s}$ chunks, ensuring that the position indices within each chunk will not exceed the chunk size $s$.
Figure~\ref{fig:model} (a) illustrates the process of segmenting a sequence of 12 tokens exceeding the pretraining length 10 into 2 chunks, with each chunk comprising $s=6<10$ tokens. Then the position indices for keys and queries are scaled within the chunk size 6. Concretely, we have 
position indices for keys $P_\mathbf{k}=[\underbrace{0, 1, 2, 3, 4, 5}_{\text{chunk 0}}, \underbrace{0, 1, 2, 3, 4, 5}_{\text{chunk 1}}]$ and $P^{\text{Intra}}_\mathbf{q} = P_\mathbf{k}$, where $P^{\text{Intra}}_\mathbf{q}$ means position indices for queries during intra-chunk attention.
To formalize, in intra-chunk attention, we adjust the position indices for queries and keys as follows: 
\begin{align}
\setlength{\jot}{-2pt} 
P^{\text{Intra}}_\mathbf{q} = P_\mathbf{k} = [0,1,\ldots, l-1]\mod s.\label{eq:intra-k}
\end{align}
For the absolute indices $i$ and $j$  within the same chunk i.e., $\lfloor i/s \rfloor = \lfloor j/s \rfloor$, satisfying $0 \leq j \leq i < l$, the element $M[i][j]$ is defined as the difference between the positional encodings of the query and the key: 
\begin{align}
    M[i][j] = P^{\text{Intra}}_\mathbf{q}[i] - P_\mathbf{k}[j].
\label{eq:intra-M}
\end{align}
When $\lfloor i/s \rfloor = \lfloor j/s \rfloor$, we calculate $M[i][j]$ follows Eq.~\ref{eq:intra-M}. The computed $M$ of the previous example where we have a sequence length of 12 and a chunk size of 6, is illustrated in Figure~\ref{fig:model} (a). The intra-chunk attention score for the interaction between the $i$-th query and the $j$-th key is then calculated as:
\begin{align}
    &\mathbf{q}_i^\top\mathbf{k}_j = f(\mathbf{q},P^{\text{Intra}}_\mathbf{q}[i])^\top f(\mathbf{k},P_\mathbf{k}[j]).
\end{align}

\subsection{Inter-Chunk Attention}
To aggregate information from other chunks, we introduce Inter-Chunk Attention.
In Llama-based LLMs, the position indices for queries are greater than those of the keys to reflect the left-to-right information flow, i.e, we have $P_\mathbf{q}[i] \geq P_\mathbf{k}[j]$ whenever $i\geq j$. Using $P_\mathbf{q}=P^{\text{Intra}}_\mathbf{q}$ and $P_\mathbf{k}$ for attention calculation between different chunks clearly violates this property. For example, considering $\mathbf{q}_s$ and $\mathbf{k}_1$  where $s$ is the chunk size, their relative distance given by $P^{\text{Intra}}_\mathbf{q}[s] = 0$ and $P_\mathbf{k}[1]=1$ is -1.
We maintain the position indices for keys $P_\mathbf{k}$ considering KV cache and seek for a new set of $P_\mathbf{q}$ during inter-chunk attention, noted as $P^{\text{inter}}_\mathbf{q}$.
Given Eq.~\ref{eq:intra-k}, the position indices for keys are cyclically repeated with the maximum position index $\max(P_\mathbf{k})=s-1$. To ensure that the queries have larger position indices than all keys from previous chunks,
A simple strategy to distinguish distant queries is to assign them a considerably large position index, such as the maximum position index during pretraining $c-1 > \max(P_\mathbf{k})$, where $c$ is the pretraining context length:
\begin{align}
\label{eq:inter-q}
    P^{\text{Inter}}_\mathbf{q} &= [\underbrace{c-1,c-1, \ldots c-1}_{\text{$l$ elements}}],
\end{align}
When $\lfloor i/s \rfloor \neq \lfloor j/s \rfloor$, we can give the relative position matrix $M$ with $\mathbf{q}_i$ and $\mathbf{k}_j$ from distinct chunks as:
\begin{align}
    M[i][j] = P^{\text{Intra}}_\mathbf{q}[i] - P_\mathbf{k}[j] = c-1-P_\mathbf{k}[j] \geq c-s.
\label{eq:inter-M}
\end{align}
As reflected in Figure~\ref{fig:model} (b),  we assign $P^{\text{Inter}}_\mathbf{q}$ with a constant value of $c-1=9$ for all positions, which is larger than the maximum position index $s-1=5$ in $P_\mathbf{k}$. We complete the rest part of the matrix $M$ left blank by intra-chunk attention with Eq.~\ref{eq:inter-M}. 

\subsection{Successive-Chunk Attention} 
Successive-Chunk Attention can be viewed as a special case for inter-chunk attention, proposed to maintain the locality of LLMs where locality means LLMs tend to heavily rely on the neighboring tokens to predict the next token~\cite{xiao2023efficient, han2023lminfinite}. Simply using inter-chunk attention may no longer keep the precise relative position between neighboring tokens, leading to performance degradation.

As shown in Figure~\ref{fig:model}(b), where the chunk size is $s=6$ and the pretraining length is $c=10$, the last key of the first chunk, $\mathbf{k}_5$, with $P_\mathbf{k}[5] = 5$, is followed by the first query of the second chunk, $\mathbf{q}_6$, with the position index $P^{\text{Inter}}_\mathbf{q}[6] = 9$. Despite their absolute distance being 1, the relative distance between $\mathbf{q}_6$ and $\mathbf{k}_5$ is $P^{\text{Inter}}_\mathbf{q}[6]- P_\mathbf{k}[5] = 4$. This configuration challenges the model's ability to maintain locality in its attention mechanism. 

Fortunately, this issue only occurs between successive chunks, so we introduce a new successive-chunk attention to deal with this case. Concretely, we propose to maintain the locality of $w$ neighboring tokens via adjusting the first $w$ position indices in for $P^{\text{Inter}}_\mathbf{q}$. For example, in Figure~\ref{fig:model} (c), given pretraining context $c=10$, chunk size $s=6$, and $P^{\text{Inter}}_\mathbf{q}=[\underbrace{9,9,9,9,9,9}_{\text{chunk 0}},\underbrace{9,9,9,9,9,9}_{\text{chunk 1}}]$, the position indices $P^{\text{Succ}}_\mathbf{q}$ can be set to $[\underbrace{\textcolor{blue}{6,7,8,9},9,9}_{\text{chunk 0}},\underbrace{\textcolor{blue}{6,7,8,9},9,9}_{\text{chunk 1}}]$ for attention calculation between successive chunks, if we keep a local window of $w=4$.
Formally, given chunk size $s$, pretraining size $c$ and local window $w$ we have:
\begin{align}
\label{eq:succ}
     P^{\text{Succ}}_\mathbf{q} = [\underbrace{\overbrace{s, s+1, \ldots, s+w-1}^{\text{$w$ elements}}, c-1, \ldots, c-1}_{\text{the same for all chunks}}],
\end{align}
where $w$ means the local window size and can be directly set to the difference between pretraining length and chunk size $c-s$. 
For $i,j$ from successive chunks, the calculation results of $M[i][j]$ using $P^{\text{Succ}}_\mathbf{q} \text{and} P_\mathbf{k}$  are reflected in Figure~\ref{fig:model} (c) where the shadow means the resulting local window.
Eq~\ref{eq:succ} ensures that the neighboring $w$ keys have the closest distance to the current query. 

By combining intra-chunk, inter-chunk, and successive-chunk attention, we finally calculate $M[i][j]$ as:
\[
M[i][j] = 
\begin{cases} 
 P^{\text{Intra}}_\mathbf{q}[i] - P_\mathbf{k}[j] & \text{if $\lfloor i/s \rfloor - \lfloor j/s \rfloor=0$}\\
 P^{\text{Succ}}_\mathbf{q}[i] - P_\mathbf{k}[j] & \text{if $\lfloor i/s \rfloor - \lfloor j/s \rfloor=1$}\\
  P^{\text{Inter}}_\mathbf{q}[i] - P_\mathbf{k}[j] & \text{if $\lfloor i/s \rfloor - \lfloor j/s \rfloor>1$}.
\end{cases}
\]

The inner product of $\mathbf{q}, \mathbf{k}$ in DCA is consequently defined as:
\begin{equation}
\small
\label{eq:dca_attn}
\mathbf{q}_i^T\mathbf{k}_j = 
\begin{cases} 
f(\mathbf{q},P^{\text{Intra}}_\mathbf{q}[i])^T f(\mathbf{k},P_\mathbf{k}[j]), & \text{if $\lfloor i/s \rfloor - \lfloor j/s\rfloor=0$}\\
f(\mathbf{q},P^{\text{Succ}}_\mathbf{q}[i])^T f(\mathbf{k},P_\mathbf{k}[j]), & \text{if $\lfloor i/s \rfloor - \lfloor j/s \rfloor=1$}\\
f(\mathbf{q},P^{\text{Inter}}_\mathbf{q}[i])^T f(\mathbf{k},P_\mathbf{k}[j]), & \text{if $\lfloor i/s \rfloor - \lfloor j/s \rfloor>1$},
\end{cases}
\end{equation}

\subsection{Normalization}
\paragraph{Softmax layer} The inner product calculations within the DCA are formalized as shown in Equation~\ref{eq:dca_attn}. Subsequently, a softmax function is applied to normalize the computed inner products:
\begin{align}
\mathbf{p}_i &= \text{softmax}(\left[ \frac{\mathbf{q}_i^\top\mathbf{k}_0}{\sqrt{d}}, \frac{\mathbf{q}_i^\top\mathbf{k}_1}{\sqrt{d}}, \ldots, \frac{\mathbf{q}i^\top\mathbf{k}_{i}}{\sqrt{d}}\right]).
\end{align}
where $d$ denotes the dimension of hidden states.

\paragraph{Flash Attention}
The PyTorch-style pseudocode for how integrating DCA with Flash Attention 2~\cite{dao2023flashattention2},  can be found in Algorithm~\ref{alg:DCA-detailed}. The explanation and complexity analysis of the code can be found in Appendix~\S\ref{sec:flash}. With Flash Attention, DCA attains comparable GPU memory usage and inference speed to the original self-attention in Llama. Results can be found in Figure~\ref{fig:effi}.

\section{Experiments}
\label{sec:exp}
We evaluate our framework, DCA, on various variants of Llama2~\cite{llama2}, specifically the 7B, 13B, and 70B models, along with their chat counterparts, which have a 4k pretraining context. Our Llama2-based model is denoted as \textbf{\model}. Additionally, we apply DCA to two popular open-source long context models: (1) Together-32k~\cite{together}\footnote{\url{https://huggingface.co/togethercomputer/LLaMA-2-7B-32K}}: This model uses Positional Interpolation (PI) as its positional encoding. The DCA-enhanced version of this model is referred to as ChunkTogether.
(2) CodeLlama~\cite{codellama}\footnote{\url{https://huggingface.co/codellama}}: This model applies NTK-Aware RoPE. Following the application of DCA, the resulting model is termed ChunkCodeLlama. 
\subsection{Experimental Setup}
DCA can be implemented by a monkey patch to replace the inference code of the original \texttt{LlamaAttention}.
Thanks to Flash Attention 2~\cite{dao2023flashattention2}, for the 7B/13B variants of \model, we only need one single NVIDIA A100-80G GPU for the inference. When scaling up to 70B models, two A100 GPUs are enough to manage inference within a 16k context length. 
The chunk size $s$ can be typically set to $\frac{3}{4}$ training length and for Llama2, this value is 3072. The number of chunks depends on the input sequence length.

In addition to training-free evaluations, we also provide finetuned models from 7B/13B Llama2 checkpoints. This finetuning process leverages only long conversations with 16k input tokens, following Vicuna~\cite{vicuna} and LongChat~\cite{longchat}. The training dataset is sourced from ShareGPT\footnote{\url{https://sharegpt.com/}} and AlpacaGPT4~\cite{alpaca}. For the data derived from ShareGPT, we specifically curate a subset by extracting responses generated by GPT-4, and dialogues that exceed 4k tokens in length. This selection results in a compilation of 5,405 training instances.

We adhere to the training hyperparameters as specified in the LongChat repository\footnote{\url{https://github.com/DachengLi1/LongChat}}. We further finetune Llama2 with over 16k steps with a batch size of 1. The finetuning process amounts to approximately 40 GPU hours for the 7B model and 60 GPU hours for the 13B variant.

\vspace{-1.0em}
\paragraph{Datasets}
We evaluate the long sequence language modeling performance of our \model on the book corpus dataset PG19~\cite{pg19}, with context lengths ranging from 4k to 192k tokens. For the 7B and 13B models, we employ a sliding window of 256, in line with previous work~\cite{yarn, longlora}. For 70B models, we adjust the sliding window size to 2048 and when dealing with contexts that exceed 96k tokens, we adjust the sliding window to be half of the input length considering the running time.
For few-shot experiments, we follow the settings in Llama2 Long~\cite{llama2long}. Concretely, we evaluate 0-shot performance of \model on NarrativeQA~\citep{narrativeqa},  1-shot on QMSum~\citep{qmsum},  2-shot on QuALITY~\citep{quality} , and 2-shot for Qasper~\citep{qasper}.
For zero-shot experiments, we test \model on 4 closed-ended tasks from L-Eval~\cite{leval}: TOFEL, QuALITY (cleaned from \citet{quality}), Coursera, SFiction. We also validate our model on passkey retrieval used in \citet{passkey}.  Evaluations on passkey retrieval~\cite{passkey} can be found in Appendix~\ref{sec:passkey}.

\vspace{-1.0em}
\paragraph{Baselines}
We compare with popular open-source long-context models available in Huggingface Transformers\footnote{prior to December 1, 2023}. \textit{Base Models}: Focused Transformer 3B~\citep{focused}, CLEX 7B~\cite{chen2023clex}, YaRN 7B/13B~\citep{yarn}, MPT 30B~\citep{MPT7b,MPT30b}, Together 7B~\citep{together}, CodeLlama 7B~\cite{codellama}, Longlora 13B/70B~\cite{longlora}, and Llama2 Long 7B/13B/70B~\cite{llama2long}.
\textit{Chat Models}:  LongChat-v1.5-32k 7B~\cite{longchat}, Vicuna-v1.5-16k~\citep{vicuna} 7B/13B, Longlora-Chat 70B~\cite{longlora}, and Llama2 Long-Chat 70B~\cite{llama2long}.

\subsection{Long-Sequence Language Modeling}\label{exps_language_modeling}
Table \ref{tab:ppl_pg19} presents the Perplexity (PPL) scores on the PG19 validation set for various training-free and finetuned models. All these baselines are Llama-based. We demonstrate that the previously best training-free method fails with a context length of 16k. However, \model can extrapolate to a context window of more than 32k, with only an increase of 0.02 in PPL.
We further demonstrate that \model surpasses the results of finetuned models within a 16k context length. Notably, the 70B variant of \model exhibits consistency in performance across a range of context lengths, achieving a PPL score that only marginally rises from 5.18 to 5.59.

We also reveal that DCA can be integrated with models that have been further trained on longer contexts with PI~\cite{pi} or NTK-Aware RoPE~\cite{fixedNTK, dynamicNTK} and support a context length of 192k in Table~\ref{tab:ppl_lclm}.
The encouraging outcomes observed with 64k input tokens motivate us to test \model on even longer contexts. We progressively tested the model with input token lengths extending from 32k to 192k (Table~\ref{tab:ppl_lclm}). For Llama2 70B, DCA has proven effective in extending the context window to 96k tokens. This extension is achieved with only a minor increase of 0.56 PPL compared to its original performance at a 4k context length.  Alongside evaluating \model, we also applied DCA to existing long-context models that utilize different positional encodings.
Integrating DCA with existing long-context models requires only an adjustment of the chunk size within the DCA framework. We show that CodeLlama and Together's Llama2 fork can be efficiently scaled to a 192k context length using DCA with a chunk size of 24k. We further validated the performance of our model on the passkey retrieval task~\cite{passkey}. The results also indicate that by integrating DCA with existing long-context models, the enhanced system maintains a 90\% retrieval accuracy with an extended context length of up to 192k tokens (Figure~\ref{fig:passkey_long}).

\begin{table}[t!]
\vspace{-0.5em}
    \caption{Perplexity (PPL) evaluation on PG19~\cite{pg19} validation set. The results highlighted in \textcolor{red}{red} indicate the Perplexity has increased by more than 1.0 compared to its original value at the pretraining context length of 4096. ReRoPE~\cite{rerope2023} encounters  OOM (Out of Memory) problems with 16k input tokens as it is currently not compatible with Flash Attention. The scaling factors in PI and NTK are dynamically changed.}
    \label{tab:ppl_pg19}
    \small
    \centering
    \begin{adjustbox}{max width=1.00\linewidth, center}
    \begin{tabular}{l|ccccc}
        \toprule
        \multirow{2}{*}{\textbf{Model}} & \multicolumn{5}{c}{Evaluation Context Window} \\
            & 4096 & 8192 & 16384 & 32768 & 65536  \\
        \midrule
        
        \multicolumn{6}{l}{\hspace{-1.5mm} {\textbf{Finetuned}} \vspace{-0.1em} \vrule width 0pt height 8pt depth 5pt} \\
        
        Longlora-32k 7B&  \bf8.14 & \bf7.85 & \bf7.70 & 7.80 & \textcolor{red}{91.79}\\
        Together-32k 7B&  8.21 & 7.95 & 7.76 & \bf7.64 & \textcolor{red}{\textgreater$10^2$}\\
        CodeLlama-16k 7B&  8.93 & 8.64 & 8.44 & 8.36 & \bf8.65\\
        CLEX-16k 7B&  8.84 & 7.66 & 7.43 & 7.57 & 8.73\\
        \midrule
        \multicolumn{6}{l}{\hspace{-1.5mm} {\textbf{Training-free}} \vspace{-0.1em} \vrule width 0pt height 8pt depth 5pt} \\
        Llama2 7B& \bf7.87 &  \textcolor{red}{\textgreater$10^2$} & \textgreater$10^2$ & \textgreater$10^2$ & \textgreater$10^2$\\
        Llama2-ReRoPE 7B&  7.94 &  7.75 & OOM & OOM & OOM\\
        Llama2-PI 7B&  7.87 & \textcolor{red}{9.19} & 15.11 & \textgreater$10^2$ & \textgreater$10^2$ \\
        Llama2-PI-Yarn 7B&  7.87 & 8.80 & \textcolor{red}{11.75} & 42.42 & \textgreater$10^2$ \\
        Llama2-NTK 7B&  7.87 & \textcolor{red}{11.98} & 26.12 & 58.91 & \textgreater$10^2$\\
        Llama2-NTK-Yarn 7B&  7.87 & 8.06 & \textcolor{red}{9.82} & 11.74 & 41.57\\
        \model 7B &  7.87 & 7.67 & 7.64 & 7.89 & \textcolor{red}{15.87}\\
        \model 13B&  7.15 & 6.95 & 6.99 & 7.90 & \color{red}{15.14}\\
        \model 70B&  \bf5.24 & \bf5.18 & \bf5.21 & \bf5.30 & \bf5.59\\
        \midrule
        \multicolumn{6}{l}{\hspace{-1.5mm} {\textbf{Llama3}} \vspace{-0.1em} \vrule width 0pt height 8pt depth 5pt} \\
        Llama3 8B & 9.04 &  8.71 & \textcolor{red}{78.88} & \textgreater$10^2$ & \textgreater$10^2$\\
        Llama3 70B & 5.36 &  5.16 & \textcolor{red}{\textgreater$10^2$} & \textgreater$10^2$ & \textgreater$10^2$\\
        \textsc{ChunkLlama3} 8B & 9.04 &  8.71 & 8.61 & 8.62 & 8.95 \\
        \textsc{ChunkLlama3} 70B &\bf5.36 & \bf5.16 & \bf5.14 & \bf5.14 & \bf5.21 \\
        \bottomrule
    \end{tabular}
    \end{adjustbox}
\end{table}

\begin{table*}[htbp!]
\vspace{-0.5em}
    \caption{Perplexity evaluation on PG19~\cite{pg19} validation set with context lengths of up to 192k tokens.  We test DCA on Llama2 70B together with 2 popular further pretrained models using PI and NTK. The results highlighted in \textcolor{red}{red} indicate the PPL has increased by more than 1.0 compared to its original value at the pretraining context length of 4096.}
    \label{tab:ppl_lclm}
    \small
    \renewcommand\arraystretch{0.95}
    \centering
    \begin{adjustbox}{max width=0.85\linewidth, center}
    \setlength\tabcolsep{9pt} 
    \begin{tabular}{l|cc|ccccccc}
        \toprule
        \multirow{2}{*}{\textbf{Model}} &Position &Training & \multicolumn{6}{c}{Evaluation Context Window} \\
          & Emb & context & 4k & 32k & 64k & 96k & 128k & 160k & 192k  \\
        \midrule
        Llama2 7B & RoPE &4k & 7.87 & \textcolor{red}{\textgreater$10^2$}  & \textgreater$10^2$ & \textgreater$10^2$ & \textgreater$10^2$ & \textgreater$10^2$ & \textgreater$10^2$\\
        \model 7B& RoPE & 4k & 7.87 & 7.89 & {\textcolor{red}{15.87}} & 43.57 & 96.21 & \textgreater$10^2$ & \textgreater$10^2$\\
        Llama2 70B & RoPE &4k & \bf5.24 & \textcolor{red}{\textgreater$10^2$}  & \textgreater$10^2$ & \textgreater$10^2$ & \textgreater$10^2$ & \textgreater$10^2$ & \textgreater$10^2$\\
        \model 70B & RoPE &4k & \bf5.24 & \bf5.30 & \bf5.59 & \bf5.80 & \bf6.12 & \textcolor{red}{6.52}  & 7.05\\
        \midrule
        Llama3 8B & RoPE & 8k & 9.04 & \textcolor{red}{\textgreater$10^2$}  & \textgreater$10^2$ & \textgreater$10^2$ & \textgreater$10^2$ & \textgreater$10^2$ & \textgreater$10^2$\\
        \textsc{ChunkLlama3} 8B& RoPE & 8k & 9.04 & 8.61 & 8.62 & 8.95 & 9.43 & 10.04 & 10.66\\
        Llama3 70B & RoPE & 8k & \bf5.36 & \textcolor{red}{\textgreater$10^2$} & \textgreater$10^2$ & \textgreater$10^2$ & \textgreater$10^2$ & \textgreater$10^2$ & \textgreater$10^2$\\
        \textsc{ChunkLlama3} 70B& RoPE & 8k & \bf5.36 & \bf5.14 & \bf5.14 & \bf5.21 & \bf5.32 & \bf5.40 & \bf5.45\\
        \midrule
        CodeLlama 7B & NTK & 16k & 8.93 & 8.36 & 8.65 & 9.14 & 9.87 & \textcolor{red}{15.68} & 24.78\\ 
        ChunkCodeLlama 7B & NTK & 16k& 8.93 & 8.36 & 8.13 & 8.33 & 8.66 & 9.30 & 9.83\\
        Together 7B &  PI & 32k & 8.21 & \bf7.64 & \textcolor{red}{\textgreater$10^2$}  & \textgreater$10^2$ &\textgreater$10^2$ & \textgreater$10^2$ & \textgreater$10^2$\\
        ChunkTogether 7B &  PI & 32k & 8.21 & \bf7.64 & \bf7.59 & \bf7.64 & \bf7.67 & \bf7.74 & \bf7.83 \\
        \bottomrule
    \end{tabular}
    \end{adjustbox}
\end{table*}

\begin{table*}[h!]
\vspace{-1.5em}
  \caption{ Comparison between popular open-source base models (first block) and proprietary models (last block) across four research benchmarks on their validation set.  We \underline{underline} the best results in each block. Results exceeding the previous best open-source finetuned model are in \textbf{bold}. Llama2 Long has been trained with a total of 400B tokens over 100,000 steps.
  The maximum allowed prompt length is set to 16,384 tokens.
 $^\dagger$: results are taken from~\citet{llama2long}
We use the simplest prompt: \textit{long-document Question:... Answer:}. In-context examples are randomly selected from the training set, and we also have a discussion on the selection of in-context examples in Appendix~\S\ref{sec:selection}.
  }
  \label{tab:few_shot}
  \small
  \centering
\renewcommand\arraystretch{0.95}
  \begin{adjustbox}{max width=0.82\linewidth, center}
      \begin{tabular}{l|cc|ccccc}
        \toprule
        \multirow{2}{*}{\textbf{Model}} & Further & Training & \bf NarrativeQA  & \bf Qasper  &\bf QuALITY & \bf QMSum & \multirow{2}{*}{\textbf{Avg}}\\
        & training &context & F1 (0-shot) & F1 (2-shot) & EM (2-shot) & R-g (1-shot) &  \\
        \midrule
        FoT 3B$^\dagger$ & \cmark & 8k & 16.3 & 15.4 & 20.5 & 10.6 & 15.7 \\
        Yarn 7B$^\dagger$ & \cmark & 128k  &  20.9 & 26.2 & 32.3 & 11.4 & 22.7  \\
        Together 7B$^\dagger$ & \cmark & 32k & 23.3 & 27.3 & 41.2 & 12.6 & 26.1\\
        Yarn 13B $^\dagger$& \cmark & 128k  & 23.4 & 27.1 & 46.4 & 11.9 & 27.2 \\
        Longlora 13B & \cmark & 32k & 25.8 & 26.4 & 48.9 & 15.1 & 29.1 \\ 
        MPT 30B$^\dagger$ & \cmark  & 8k & 22.9 & \underline{29.0} & 41.5 & 10.3 & 25.9 \\
        Llama2-DynNTK 70B & \color{red}{\xmark} & {4k} & 11.1 & 27.8 & 60.9 & 7.8 & 26.9 \\
        Llama2 70B$^\dagger$ & \color{red}{\xmark} & {4k} & 25.7 & 27.5 & 53.0 & 11.9 & 29.5 \\
        Longlora 70B & \cmark & 32k & \underline{34.2} & \underline{29.0} & \underline{69.9} & \underline{15.6} & \underline{37.2} \\
        \midrule
        \model 7B  & \color{red}{\xmark} & 4k  & 20.0 & 28.2 & 35.6 & 14.7 & 24.6 \\ 
        \model 13B  & \color{red}{\xmark} & {4k}  & {26.3} & \bf{29.3} & {47.9} & 15.2 & 29.7 \\ 
        \model 70B & \color{red}{\xmark} & {4k} & {32.5} & {\bf29.6} & \bf73.2 & {\bf16.0} & {\bf37.8}\\ 
        \textsc{ChunkLlama3} 8B  & \color{red}{\xmark} & {8k} &27.4 & 30.5 & 52.6 & 15.4 & 31.5\\ 
        \textsc{ChunkLlama3} 70B & \color{red}{\xmark} & {8k} & \underline{\bf33.7} & \underline{\bf33.1} & \underline{\bf75.4} & \underline{\bf16.0} & \underline{\bf39.5} \\ 
        \midrule
        \multicolumn{8}{l}{\hspace{-1.5mm} {\textit{ proprietary models}} \vspace{-0.1em} \vrule width 0pt height 8pt depth 5pt} \\
        Llama2 Long 7B$^\dagger$ & \cmark & 32k & 21.9 & 27.8 & 43.2 & 14.9 & 27.0 \\ 
        Llama2 Long 13B$^\dagger$ & \cmark & 32k & 25.6 & 31.2 & 57.6 & 15.7 & 32.5 \\ 
        Llama2 Long 70B$^\dagger$ & \cmark & 16k & \underline{30.9} & \underline{35.7} & \underline{79.7} & \underline{16.5} & \underline{40.7} \\
        \bottomrule
      \end{tabular}
  \end{adjustbox}
\end{table*}

\subsection{Practical Tasks}
In contrast to previous studies that typically validate their methods based on PPL, we also apply our framework to both base models and instruction-finetuned chat models on real-world benchmarks.

\paragraph{Few-shot Results}
We validate DCA on models that have not undergone instruction tuning in a few-shot learning setting. The results are summarized in Table \ref{tab:few_shot}.  Experimental settings are the same as those in~\citet{llama2long}. If the input prompts exceed an input length of 
 16k tokens, they are truncated from the left side.
Most test cases within NarrativeQA~\cite{narrativeqa} and QMSum~\cite{qmsum} have input lengths exceeding 16k tokens, while the lengths of test cases in Qasper~\cite{qasper} and QuALITY~\cite{quality} are generally under 8k tokens. 
Without any training cost, both the 7B/13B variants of \model achieve results comparable to popular finetuned baselines such as YaRN~\cite{yarn}, MPT~\cite{MPT7b}, Together~\cite{together}, which are based on previous scaled RoPE~\cite{pi, fixedNTK} or Alibi~\cite{alibi}. 

Unlike previous studies that usually verify their techniques on smaller versions of Llama2, we also present results for DCA paired with Llama2 70B, where DCA improves performance by an average of more than 8.0 points over the original Llama2 model with a 4k training length. 
Given the increasing cost of long-context finetuning for 70B models, we did not find many open-source 70B baselines. We compare our training-free method against the robust 70B baseline, Longlora~\cite{longlora}, which employs LoRA-based~\cite{hu2021lora} efficient tuning based on the Redpajama dataset~\cite{together2023redpajama} for 1000 steps supporting a 32k context window. The results demonstrate that our 70B DCA model achieves comparable performance (37.8 vs. 37.2) requires no training steps.

Compared to the strong proprietary baseline, Llama2 Long~\cite{llama2long}, which has been trained with a total of 400 billion tokens (Llama2 pretraining corpus and new long text data) over 100,000 steps, the performance gaps for all sizes of models are generally within a 3-point range. The in-context examples used in this experiment are randomly selected from the training set. We have also tried other ways to select the examples, and the details are included in Appendix \ref{sec:selection}.

\begin{table*}[t!]
\vspace{-0.5em}
\caption{Comparison with open-source \textbf{chat} models (first block) and proprietary models (last block) on 4 closed-ended tasks with various input lengths from L-Eval~\cite{leval}. We \underline{underline} the best results in each block. Results exceeding previous the best open-source finetuned model are in \textbf{bold}. `dialogues' means the mix of ShareGPT and AlpacaGPT4 used in our training. Llama2-PI-SFT and Llama2-NTK-SFT are models trained with the same data and training steps with \model. $^\ddagger$: results are taken from~\citet{llama2long}.
}
  \label{tab:zero_shot}
  \small
  \centering
  \begin{adjustbox}{max width=0.75\linewidth, center}
      \begin{tabular}{l|cc|ccccc}
        \toprule
        \multirow{2}{*}{\textbf{Model}} & Finetuning & Training & \bf TOFEL  & \bf QuALITY  &\bf Coursera & \bf SFiction & \multirow{2}{*}{\textbf{Avg}}\\
        & corpus & context & (3k$\sim$5k) &  (4k$\sim$9k) &  (5k$\sim$17k) &  (6k$\sim$27k) &  \\
        \midrule
        Llama2-Chat 7B & \color{red}{\xmark} & 4k  & 51.67 & 37.62 & 29.21 & 60.15 & 48.74\\
        Llama2-DynNTK 7B & \color{red}{\xmark} & 4k  & 52.27 & 30.69 & 13.95 & 57.02 & 38.48\\
        Longchat-v1.5-32k 7B  & ShareGPT & 32k & 39.77 & 37.62 & 32.99 & 57.02 & 41.85\\
        Llama2-PI-SFT 7B& Dialogues & 16k &  56.13 & 38.61 & 36.19 & 53.90 & 46.20\\
        Llama2-NTK-SFT 7B& Dialogues & 16k & 53.90 & 38.11 & 34.01 & 64.06 & 47.51\\
        Vicuna-v1.5-16k 7B & ShareGPT & 16k & 55.39 & 39.60 & 38.66 & 60.15 & 48.45\\
        Llama2-Chat 13B & \color{red}{\xmark} & 4k  & 60.96 & 42.57 & 35.75 & 54.68 & 48.99\\
        Llama2-DynNTK 13B & \color{red}{\xmark} & 4k & 62.45 & 33.16 & 37.06 & 60.93 & 48.40\\
        Vicuna-v1.5-16k 13B & ShareGPT & 16k & 68.40 & 53.96 & 40.69 & 61.71 & 56.19\\
        Longlora-Chat 70B &  LongAlpaca & 32k & \underline{71.37} & \underline{55.45} & \underline{44.76} & \underline{67.96} & \underline{59.88}\\
        \midrule

        \multicolumn{8}{l}{\hspace{-1.5mm} {\textbf{Training-free}} \vrule width 0pt height 8pt depth 5pt} \\
        \model-Chat 7B & \color{red}{\xmark} & 4k & {57.62} & 
        {35.14} & 32.12 & {61.72} & {46.64}\\
        \model-Chat 13B & \color{red}{\xmark} & 4k & {66.54} & {43.06} & {41.56} & {57.03} & {52.04}\\
        \model-Chat 70B & \color{red}{\xmark} & 4k & \underline{\bf82.15} & \bf60.39 & \bf48.54 & {61.72} & \bf63.20\\
        
        \multicolumn{8}{l}{\hspace{-1.5mm} {\textbf{Llama3}} \vrule width 0pt height 8pt depth 5pt} \\
        \textsc{ChunkLlama3}-Instruct 8B &  \color{red}{\xmark} & 8k & 83.27 & 63.86 & 56.24 & 70.31 & 68.42\\ 
        \textsc{ChunkLlama3}-Instruct 70B &  \color{red}{\xmark} & 8k &\underline{84.75} & \underline{82.17} & \underline{76.88} & \underline{75.78} & \underline{79.89}\\
        
        \multicolumn{8}{l}{\hspace{-1.5mm} {\textbf{Finetuned}} \vrule width 0pt height 8pt depth 5pt} \\
        \model-Chat 7B & Dialogues & 16k & 62.08 & 41.58 & 39.68 & 64.06 & 51.85\\ 
        \model-Chat 13B & Dialogues & 16k & 65.42 & 53.96 & \underline{44.76} & \underline{65.62} & 57.94\\
        \midrule
        \multicolumn{8}{l}{\hspace{-1.5mm} {\textit{ proprietary models}} \vspace{-0.1em} \vrule width 0pt height 8pt depth 5pt} \\
        GPT3.5-16k-0613 & Unkown & -- & 78.43 & \bf61.38 & 63.51 & 64.84 & 67.03\\
        Claude1.3-100k & Unkown & -- &\bf83.64 & 60.03 & \bf73.76 & \bf72.65 & \bf72.52 \\
        Llama2 Long-Chat 70B$^\ddagger$ & Long doc+diag & 16k  &  81.8 & -- & 52.9 & -- & --\\

        \bottomrule
      \end{tabular}
  \end{adjustbox}
\end{table*}

\paragraph{Zero-shot Results}
In addition to verifying DCA on base models, we also apply DCA on the \textbf{chat version} of Llama2 (with instruction tuning) in a zero-shot learning scenario. Specifically, we test our models on four closed-ended tasks from L-Eval~\cite{leval} with diverse input lengths ranging from 3k to 27k. All these datasets adopt Exact Match (EM) as the evaluation metric. Overall, the conclusions are similar to the few-shot evaluation. Our training-free 7B/13B models show comparable performance with open-source models with further training.
Notably, in zero-shot experiments, we demonstrate a significant improvement over the Chat version of Longlora 70B~\cite{longlora}.  
Furthermore, when compared with proprietary models such as GPT-3.5 with a 16k token context and the chat version of Llama2 Long, the results suggest that the Llama2 70B chat model can be directly scaled to a 16k context window without additional training with DCA, achieving 94\% of the performance of \texttt{gpt-3.5-turbo-16k}. We also demonstrate that our model's performance can be enhanced through additional finetuning on long dialogue data following the approach used by Vicuna \cite{vicuna} and Longchat \cite{longchat}, both of which are popular finetuned baselines utilizing ShareGPT. With further training, \model-Chat outperforms the previously best 13B model, Vicuna-v1.5-13b-16k, by a significant margin of 1.75 points.

\subsection{Analysis}
\begin{figure}[htbp!]
\centering
    \includegraphics[width=0.47\textwidth]{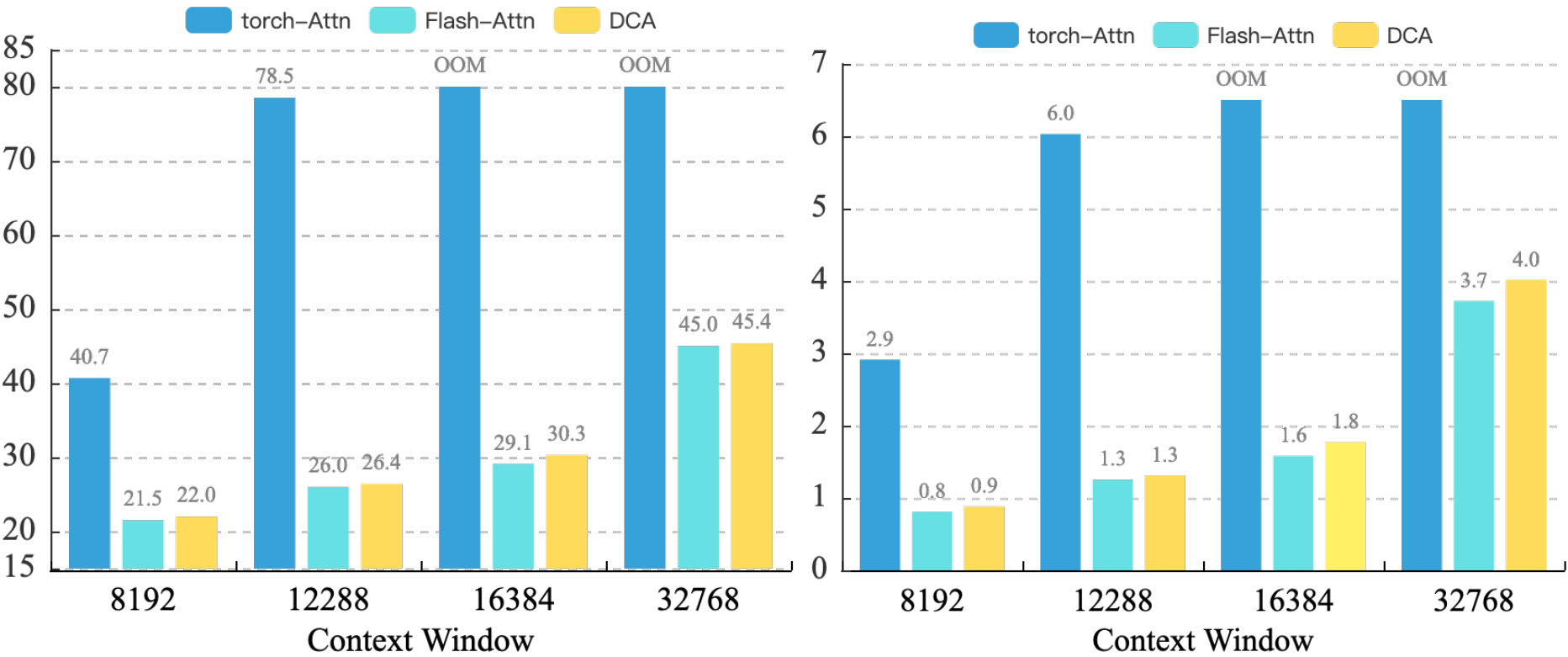}
   \vspace{-1.2em}
  \caption{Inference time and GPU memory of (a) the original self-attention implemented by Pytorch, (b) Flash Attention~\cite{dao2023flashattention2}, and (c) DCA (this work).}
  \label{fig:effi}
\end{figure}

\paragraph{Efficiency}
In figure~\ref{fig:effi}, the inference time and GPU memory of (a) the original self-attention mechanism as implemented in PyTorch, Flash Attention~\cite{dao2023flashattention2}, and our proposed DCA (integrated with Flash Attention) are evaluated across various prompt lengths. These experiments are run on a single NVIDIA 80G A100 GPU using Llama2 7B. The input long prompt is from NarrativeQA~\cite{narrativeqa}. We conduct 20 trials and report the average performance. Without Flash Attention, we observe that the maximum input length manageable by a single GPU is roughly between 12k and 16k tokens. DCA sustains similar GPU memory consumption and inference speed, without adding considerable overhead, with the original Flash attention.

\vspace{-0.5em}

\begin{figure}[htbp!]
\centering
    \includegraphics[width=0.49\textwidth]{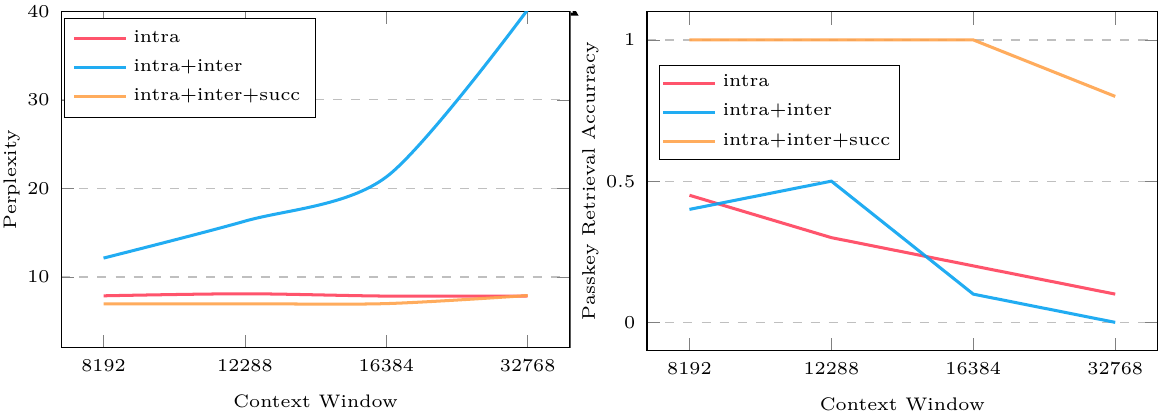}
   \vspace{-1.5em}
  \caption{Ablation study of DCA on language modeling (left) and passkey retrieval (right). We test the three attention mechanisms with input sequences from 8k to 32k.}
  \label{fig:ablation}
\end{figure}

\paragraph{Ablation Study}\label{sec:ablation}
To validate the three attention mechanisms proposed in this work, we present an ablation study for DCA in Figure~\ref{fig:ablation}, focusing on language modeling and passkey retrieval tasks. We consider three experimental conditions:
(1) Employing only intra-chunk attention.
(2) Utilizing both intra-chunk and inter-chunk attention.
(3) Combining all three types of attention: intra-chunk, inter-chunk, and successive chunk attention.
From the results in language modeling, we observe that using intra-chunk attention which disregards information from previous chunks, is able to maintain a very low PPL but hinders the model's ability to retrieve passkeys from other chunks. Introducing inter-chunk attention, we notice an improvement in passkey retrieval performance at an input length of 12k. However, the loss of locality causes a significant increase in the model's PPL. By integrating successive chunk attention, we achieve both a low PPL and high retrieval accuracy.

\vspace{-0.5em}

\section{Conclusion}
In this paper, we present Dual Chunk Attention (DCA) as a novel and efficient approach to overcoming the context length limitations inherent in LLMs. By ingeniously leveraging the model's existing position indices and introducing a multi-faceted attention mechanism, DCA allows for extrapolating more than 8x the training length without the need for costly and time-consuming further training.

\section*{Impact Statement}
Numerous studies have emerged targeting to expand the supported context length of LLMs; however, due to high training costs and incompatibilities with technologies such as Flash Attention, the industry mainly relies predominantly on expanding the base frequency of RoPE or PI. Our Dual Chunk Attention (DCA) method is compatible with Flash Attention and requires only modifications to the inference code, negating the need for extensive retraining. DCA preserves model performance within the training length, and only benefits it beyond this range, offering compatibility with models that have already undergone long-context finetuning. Consequently, our approach may have a substantial impact on the industry, providing a cost-effective solution for managing long-context scenarios in LLM applications.
There are many potential societal consequences of our work, none of which we feel must be specifically highlighted here.

\section*{Acknowledgements}
We thank Yukang Chen and Hang Yan for their helpful comments and open-source code. This research was supported in part by the joint research scheme of the National Natural Science Foundation of China (NSFC) and the Research Grants Council (RGC) under grant number
N\_HKU714/21.

\bibliography{chunkllama}
\bibliographystyle{icml2024}

\newpage
\appendix
\onecolumn
\section{Appendix}
\subsection{Passkey retrieval}\label{sec:passkey}
\begin{figure*}[htbp!]
\centering
    \includegraphics[width=0.95\textwidth]{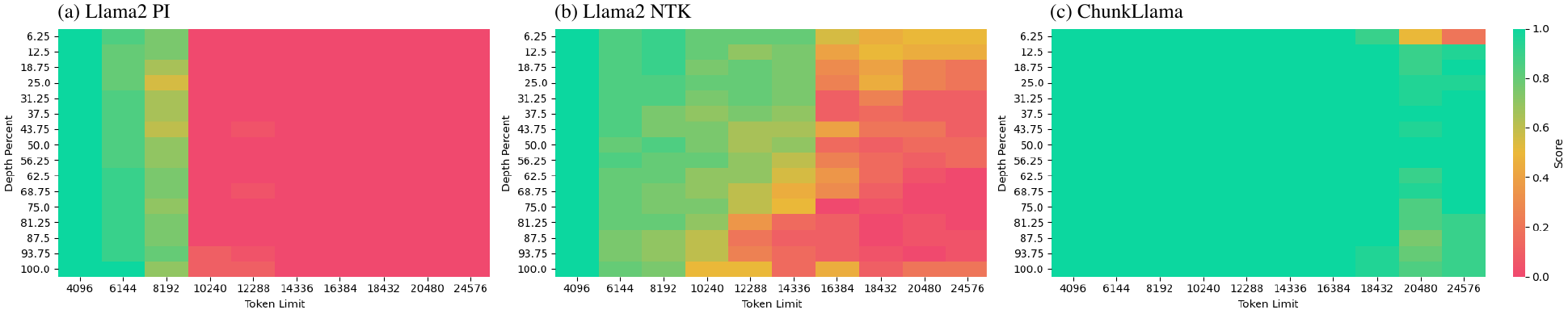}
\vspace{-0.5em}
  \caption{ Testing Different Learning-Free Extension Methods with a 24K Context ("Needle in a Haystack" Passkey Retrieval). All the models have a 4k pretraining context and are not further trained. The X-axis represents the input context length, and the Y-axis indicates the depth of the passkey within the document. For each depth, we run 20 different test cases.
  }
  \label{fig:passkey}
\end{figure*}

In addition to practical tasks, we evaluate the long-context capability of LLMs to perform the passkey retrieval task as defined in~\citet{passkey}. This task challenges a language model to locate a simple passkey (e.g., a five-digit random number) embedded within a lengthy and otherwise nonsensical text sequence. The primary purpose of this task is to determine if a Large Language Model (LLM) can maintain awareness of information distributed throughout a lengthy input sequence. To assess retrieval accuracy, we randomly place the passkey at various document depths which are distributed uniformly. For each document depth, we run 20 times with different passkeys and we test the input sequence length from 4k to 20k. We compare the performance of DCA with 2 popular extension methods: PI~\cite{pi}, NTK-Aware~\cite{fixedNTK, dynamicNTK}, on the Llama2 13B model with a 4k pretraining context window. The performance results are depicted in Figure~\ref{fig:passkey}. Notably, within a context length of 18k tokens, our model \model consistently achieved a 100\% passkey retrieval accuracy across all depths tested.

\begin{figure}[htbp!]
\centering
    \includegraphics[width=0.8\textwidth]{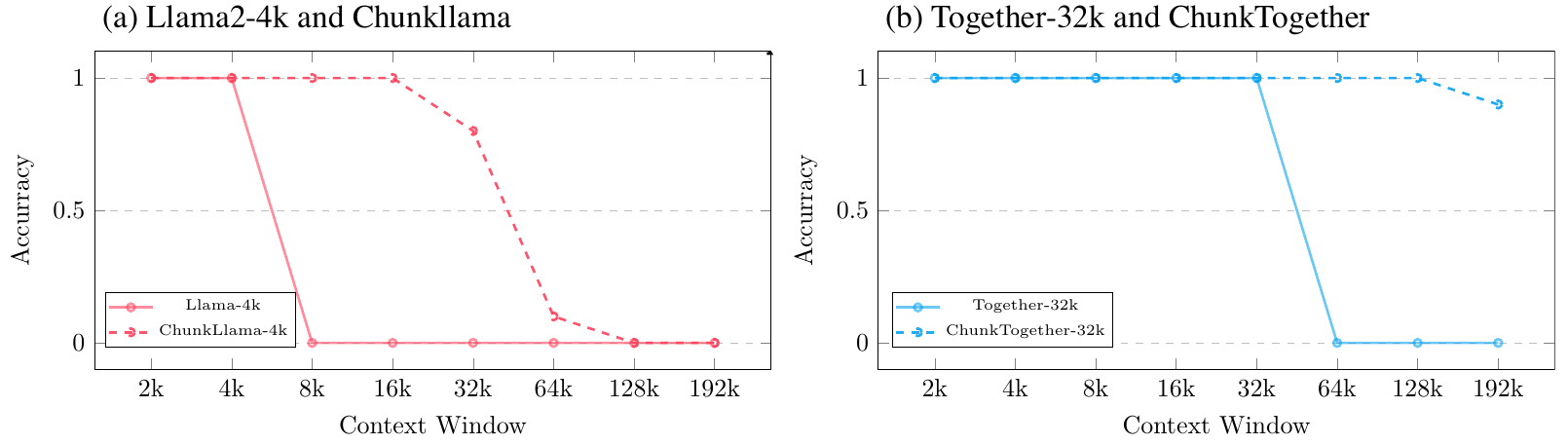}
   \vspace{-1.2em}
  \caption{Passkey retrieval over a 192k context length for Llama2 13B, Together-32k 7B and their DCA enhanced versions.}
  \label{fig:passkey_long}
\end{figure}

We expanded the scope of the passkey retrieval tasks by incrementally increasing the input token count from 2k to 192k. For each input context length, the model is evaluated 20 times, with the passkey's position randomly varied in each test. Additionally, we also verify the Together-32k 7B model~\cite{together}, which supports a 32k token context window, and its ChunkTogether 7B counterpart.  The outcomes for both the baseline and DCA-enhanced variants of these models are illustrated in Figure~\ref{fig:passkey_long}. With only a 4k training context length, \model maintains high retrieval accuracy up to a 32k context length. By integrating these findings with existing long-context models, we can feasibly extend the supported context window to an impressive 192k tokens using a learning-free approach.

\textit{lost in the beginning}: An intriguing observation is that the failure cases of PI appear to be largely unrelated to the document's depth, while the NTK-based approach typically excels when the passkey is positioned near the beginning of the document. However, its effectiveness significantly diminishes—with accuracy dropping to between 40\% and 80\%—when the passkey is placed in the middle sections. This trend aligns with findings reported by \citet{lost_in_the_middle}. Conversely, as the input context is expanded, \model demonstrates improved performance in the middle sections but the first place where a drop in accuracy occurs is at the beginning of the text.

\begin{figure}[htbp!]
\centering
    \includegraphics[width=0.8\textwidth]{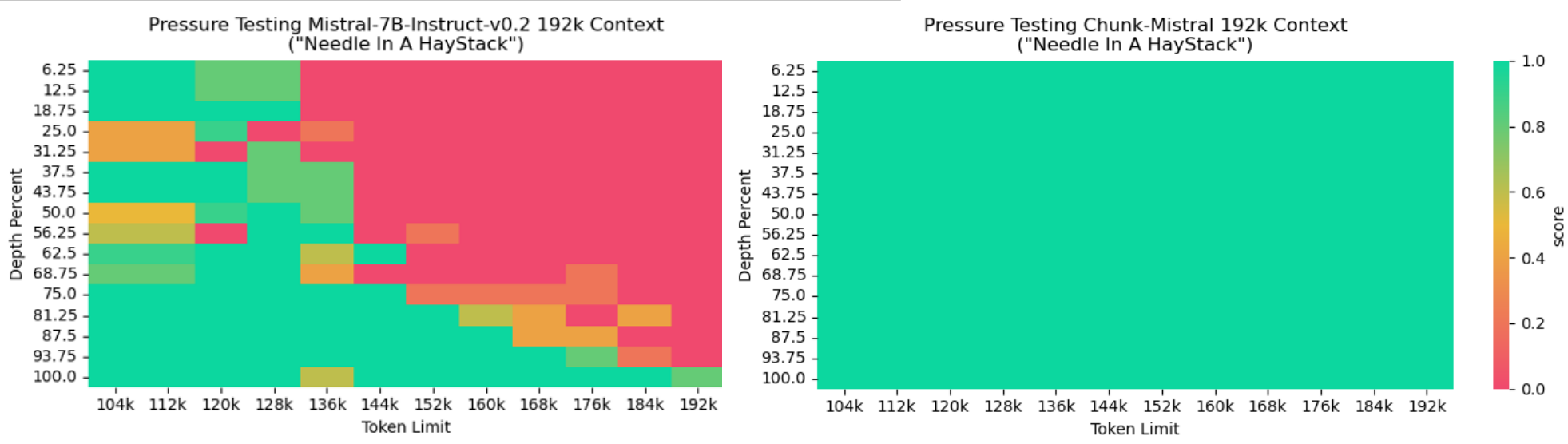}
   \vspace{-1.2em}
  \caption{Pressure testing Mistral-7B-Instruct-v0.2 over a 192k context length for and its DCA enhanced version (("Needle In A HayStack")). }
  \label{fig:mistral}
\end{figure}

\begin{figure}[htbp!]
\centering
    \includegraphics[width=0.8\textwidth]{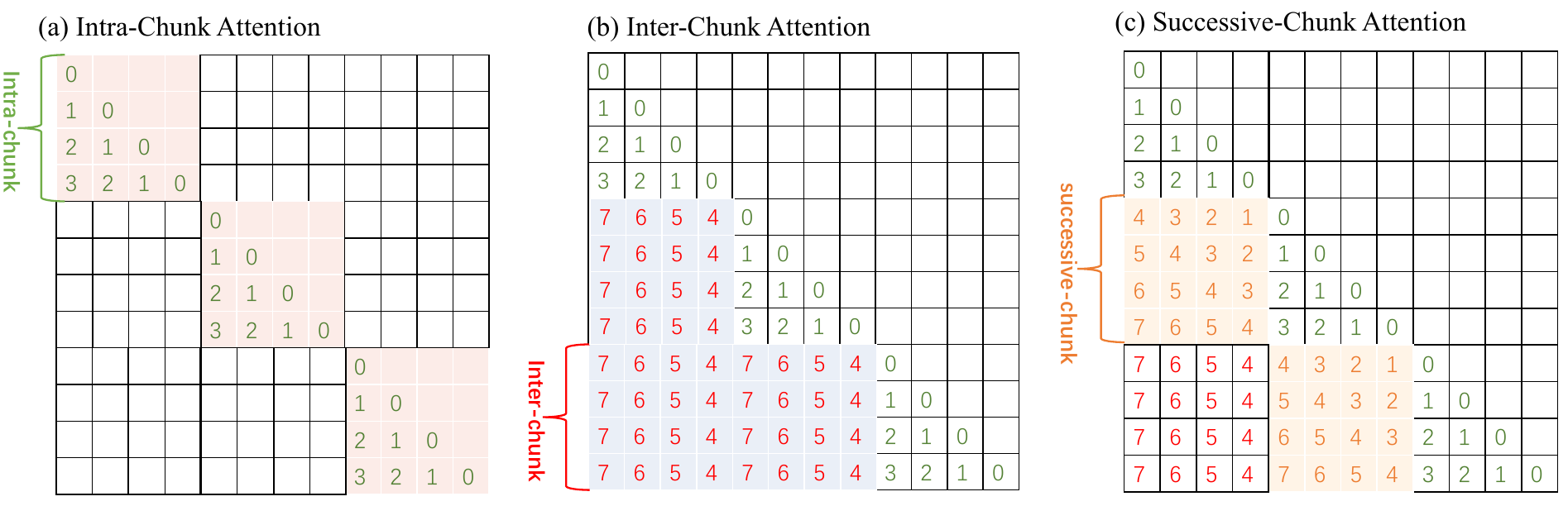}
   \vspace{-1.2em}
  \caption{{Visualization of the Relative Position Matrix $M$ employing Dual Chunk Attention (DCA) by splitting the whole sequence into \textbf{3 chunks} and the chunk size $s=4$. In this case, we have the pretraining window size $c=8$, and local window size $w=3$. The sequence is segmented into 3 chunks to ensure that relative positions do not exceed 7. The matrix element $M[i][j]$ represents the relative position between the $i$-th query vector $\mathbf{q}$ and the $j$-th key vector $\mathbf{k}$.}}
  \label{fig:illu}
\end{figure}
\subsection{{More Examples}}
In this section, we give an example of handling a 12-token sequence but the pre-training length is only 8. 
Based on Llama2, the key/query position indices will be initialized as:
\begin{align*}
P_\mathbf{q} &= [0, 1, 2, 3, 4, 5, 6, 7, 8, 9, 10, 11] \\
P_\mathbf{k} &= [0, 1, 2, 3, 4, 5, 6, 7, 8, 9, 10, 11].
\end{align*}
When $\mathbf{query}[11]$ and $\mathbf{query}[0]$ perform an inner product, their relative position is 11, which exceeds the pre-training size.

In DCA, we set a chunk size, which is a hyperparameter smaller than the pre-training length. In this example, we can set the chunk size to 4, which means we split the entire input into 3 chunks and give new position indices for keys:
\[
[0, 1, 2, 3, 4, 5, 6, 7, 8, 9, 10, 11] \Rightarrow [0, 1, 2, 3, 0, 1, 2, 3, 0, 1, 2, 3]
\]

We then give different position indices for queries.

\textbf{Intra-chunk Attention:} Calculating the attention for tokens in the same chunk
\begin{align*}
{P}_\mathbf{q}^{\text{intra}} &= [0, 1, 2, 3, 0, 1, 2, 3, 0, 1, 2, 3] \\
P_\mathbf{k} &= [0, 1, 2, 3, 0, 1, 2, 3, 0, 1, 2, 3]
\end{align*}
The resulting position matrix is shown in Appendix Figure~\ref{fig:illu} (a) and the maximum relative position is $3 - 0 = 3$.

\textbf{Inter-chunk Attention:} Calculating the attention for tokens in the different chunks
When the pre-training length is 8, we have the maximum position index=7.
\begin{align*}
{P}_\mathbf{q}^{\text{inter}} &= [7, 7, 7, 7, 7, 7, 7, 7, 7, 7, 7, 7] \\
P_\mathbf{k} &= [0, 1, 2, 3, 0, 1, 2, 3, 0, 1, 2, 3]
\end{align*}
The resulting position matrix after inter-chunk attention is shown in Appendix Figure~\ref{fig:illu} (b).

\textbf{Successive-chunk Attention:} Calculating the attention for tokens in successive chunks. We change the first $w=3$ (a hyperparameter) elements in ${P}_\mathbf{q}^{\text{inter}}$:
\begin{align*}
{P}_\mathbf{q}^{\text{succ}} &= [4, 5, 6, 7, 4, 5, 6, 7, 4, 5, 6, 7] \\
P_\mathbf{k} &= [0, 1, 2, 3, 0, 1, 2, 3, 0, 1, 2, 3]
\end{align*}
The position matrix after successive-chunk attention is shown in Appendix Figure~\ref{fig:illu} (c).

\subsection{Flash Attention}\label{sec:flash}
 We divide the standard self-attention into 3 separate flash attention calculations respectively obtaining the output from intra-chunk attention, inter-chunk-attention, and successive chunk-attention.  Algorithm~\ref{alg:DCA-detailed} showcases how the 3 attention introduced in DCA integrate with Flash Attention. We illustrate with the $i$-th query vector $\mathbf{q}_i$ and it needs to calculate the inner product with all keys $\mathbf{k}_j$ with $j\leq i$. We have $n=\lfloor i/s \rfloor$ chunks before the current chunks. DCA calls 3 separate Flash Attention operations with complexity $O(i-n*s)$(intra-chunk attention), $O(s)$ (succssive-chunk attention) and $O(s*(n-1))$.

 \definecolor{customTeal}{RGB}{0, 128, 128} 
\lstset{language=Python,         
        basicstyle=\fontsize{7.0pt}{7.5pt}\ttfamily\selectfont,
        keywordstyle=\color{customTeal},    
        stringstyle=\color{customTeal},    
        commentstyle=\color{customTeal},     
        morecomment=[l][\color{magenta}]{\#},
        breaklines=true,                
        showstringspaces=false          
}
\begin{figure}[htbp!]
    \begin{minipage}{1.0\textwidth}
    \begin{algorithm}[H]
    \caption{\footnotesize Pseudocode of DCA with FlashAttention}\label{alg:DCA-detailed}
\begin{lstlisting}[language=python]
    # q: 1 x d query vector (tensor with shape [1, d])
    # i: the absolute index of q (integer)
    # K, V: i x d matrices for keys and values (tensors with shape [i, d])
    # s: chunk size (integer)
    # P_k, P_q_intra, P_q_succ, P_q_inter: position ids (lists of integers)
    n = math.floor(i/s) # Number of chunks before the current chunk
    # Apply rotary position embeddings to the entire key matrix K
    K = apply_rotary_pos_emb(K, P_k) # K is [i, d] after embedding
    
    # ------------- Intra-chunk Attention, casual=True -------------
    q_intra = apply_rotary_pos_emb(q, P_q_intra[i]) # q_intra is [1, d]
    # Select intra-chunk keys and values
    K_intra = K[s*n:i] # K_intra is [(i - s*n), d]
    V_intra = V[s*n:i] # V_intra is [(i - s*n), d]
    # Compute output and softmax attention map for intra-chunk attention
    o_intra, map_intra = Flash(q_intra, K_intra, V_intra) # o_intra is [1, d], map_intra is [1, i - s*n]
    
    # ------------- Successive-chunk Attention, casual=False -----------
    q_succ = apply_rotary_pos_emb(q, P_q_succ[i]) # q_succ is [1, d]
    # Select successive-chunk keys and values
    K_succ = K[s*(n-1):s*n] # K_succ is [s, d] 
    V_succ = V[s*(n-1):s*n] # V_succ is [s, d]
    # Compute output and softmax attention map for successive-chunk attention
    o_succ, map_succ = Flash(q_succ, K_succ, V_succ) # o_succ is [1, d], map_succ is [1, s]
    
    # ------------- Inter-chunk Attention, casual=False -----------
    q_inter = apply_rotary_pos_emb(q, P_q_inter[i]) # q_inter is [1, d]
    # Select inter-chunk keys and values
    K_inter = K[:s*(n-1)] # K_inter is [s*(n-1), d]
    V_inter = V[:s*(n-1)] # V_inter is [s*(n-1), d]
    # Compute output and softmax attention map for inter-chunk attention
    o_inter, map_inter = Flash(q_inter, K_inter, V_inter) # o_inter is [1, d], map_inter is [1, s*(n-1)]
    
    # Normalization
    # Sum the attention maps for each attention type to get normalizers
    sum_intra = map_intra.sum(-1) # sum_intra is a scalar
    sum_inter = map_inter.sum(-1) # sum_inter is  a scalar
    sum_succ = map_succ.sum(-1) # sum_succ is a scalar
    normalizer = sum_intra + sum_inter + sum_succ # normalizer is a scalar
    
    # Concatenate attention outputs and divide by normalizer
    output = (sum_intra*o_intra, sum_succ*o_succ, sum_inter*o_inter) / normalizer # output is [1, d]
    \end{lstlisting}
    \end{algorithm}
    \end{minipage}
\end{figure}

\begin{table*}[h!]
  \caption{Comparison of few-shot results using different in-context examples}
  \label{tab:icl}
  \small
  \centering
\renewcommand\arraystretch{0.95}
  \begin{adjustbox}{max width=0.82\linewidth, center}
      \begin{tabular}{l|c|ccc}
        \toprule
        \multirow{2}{*}{{Models}} & \multirow{2}{*}{\textbf{In-Context Examples}} & Qasper  & QuALITY & QMSum \\
        & & F1 (2-shot) & EM (2-shot) & R-g (1-shot)  \\
        \midrule
         \model 7B& \textsc{Example Best}& 27.3 & 33.9 & 15.0 \\ 
         \model 7B & \textsc{Example Random} & 28.2 & 35.6 & 14.7 \\
         \model 7B&\ \textsc{Example Worst}& 28.4 & 35.9 & 14.3 \\
        \midrule
         \model 13B& \textsc{Example Best}& 28.5 & 46.2 & 15.6 \\ 
         \model 13B & \textsc{Example Random} & 29.3 & {47.9} & 15.2 \\
         \model 13B&\ \textsc{Example Worst}& 29.0 & {47.5} & 15.5 \\
        \bottomrule
      \end{tabular}
  \end{adjustbox}
\end{table*}
\subsection{In-Context Examples Selection}\label{sec:selection}
We opt to select in-context examples from the training set which is a practical and common way to obtain the examples~\cite{ye2023compositional,wang2024learning}. We experimented with 2 different methods for this selection process: (1)Random Selection: randomly selecting examples from the training set. 
(2) Retrieval-Based Selection: Using the current query, we employ retrieval algorithms such as BM25~\cite{robertson2009probabilistic} to find the most relevant examples from the training set. We refer to the in-context examples with the highest retrieval score as \textsc{Example  Best} and those with the lowest as \textsc{Example Worst}. The performance of different selection approaches based on \model 7B/13B is shown in Table~\ref{tab:icl}. The performance on the summarization dataset QMSum~\cite{qmsum} generally is less likely to be influenced by prompt selection. However, on the 2 question-answering datasets,
we find that using the closest examples, paradoxically, leads to the poorest outcomes and the performance of both random selection and choosing the worst example is relatively similar. A possible explanation for this phenomenon is that when the example is highly similar, LLMs tend to copy the response given in the example which usually leads to a wrong answer.

\subsection{Performance on Unseen Data}~\label{sec:examples}
Currently, almost all benchmarks for LLMs fail to thoroughly address the potential of data contamination, which implies that the test data might have already been used in pretraining or finetuning phases. To demonstrate ChunkLlama's performance on previously unseen long-document data, we directly used the Latex code of \textbf{this paper} as a test case while omitting the title, abstract, and conclusion sections. After tokenization, the total input length is 19388. We initiate the evaluation with a set of straightforward questions that do not necessitate prior knowledge for accurate responses (see Table~\ref{tab:case1}). This was followed by a series of more challenging questions designed to assess the comprehension of the proposed DCA (refer to Table~\ref{tab:case2}).

The results indicate that, compared to the NTK, \model demonstrates a superior ability to correctly interpret instructions and provide accurate responses. However, the precision of the 13B model's answers remains suboptimal, even for relatively simple inquiries. For instance, when queried about the finetuning corpus used by DCA, it erroneously cites the Llama2 corpus rather than the correct corpora, which are ShareGPT and AlpacaGPT4.

On these elementary questions, ChunkLlama 70B exhibits a remarkably high accuracy rate. Moreover, ChunkLlama 70B shows promising outcomes on more challenging queries. It is able to articulate the rationale behind our design of the inter-chunk and successive-chunk strategies accurately. However, when we pose questions that require a global understanding of this method, it still faces difficulties.

\begin{table*}[t!]
    \small
    \centering
    \caption{QA (Easy) results of \model 13B/70B and Dynamic-NTK with Llama2 13B, using the content of this paper as input. Most of the questions do not require a deep understanding of this paper. The input prompt has 19388 tokens. Dynamic-NTK is unable to give the correct answer for all test cases. Low-quality generation results are marked with \color{red}{\xmark}.  } 
    \label{tab:case1}
    \begin{tabular}{@{}c  p{0.8\textwidth}}
     \toprule
     \multicolumn{1}{r}{\textit{Prompt}} & [INST] $<<$SYS$>>$ Please read the long paper and  answer the questions $<</$SYS$>>$ 
   1. Introduction \textbackslash n\textbackslash n The ability to comprehend and process long-context information is essential for large language models (LLMs) to cater to a wide
   range of applications effectively. These include analyzing and responding to inquiries within sizable PDFs, retaining extended dialogue history, and empowering interactive chatbots... [/INST]\\
    \cmidrule{1-2}

    \multicolumn{1}{r}{\textit{Question}} & Can you give a title for this paper\\
    \multicolumn{1}{r}{\model {70B}} & \cellcolor{gray!15} Sure, here are some potential titles for the paper: 
    
    1. "Dual Chunk Attention: A Training-Free Framework for Long-Context Language Models"
    
2. "Efficient Long-Context Language Modeling with Dual Chunk Attention"

3. "Scaling Long-Context Language Models without Training: A Dual Chunk Attention Approach"

4. "DCA: A Training-Free Method for Long-Context Language Modeling"

5. "Long-Context Language Modeling with Dual Chunk Attention: A Training...\\
    \multicolumn{1}{r}{\model 13B} &  Sure! Dual Chunk Attention: A Training-Free Framework for Long-Context Scaling of Large Language Models.
    
    This title highlights the key contributions of the paper: the proposed Dual Chunk Attention (DCA) framework, which enables long-context scaling without additional training, and its application to various tasks. The phrase "training-free" is also included to emphasize the main innovation of the paper.\\
    \multicolumn{1}{r}{Dynamic NTK 13B}  & \cellcolor{gray!15} The proposed Dual Chunk Attention (DCA) framework improves the long-context scaling of Lama2 by more than 8 times, achieving a PPL of 5.59 at a context window of 2k. \color{red}{\xmark (ignore the instruction)}
 \\\cmidrule{1-2}
    \multicolumn{1}{r}{\textit{Question}} & Which model can be scaled up to 192k context with DCA? \\
    \multicolumn{1}{r}{\model {70B}} &  \cellcolor{gray!15} The Llama2 70B model can be scaled up to a 192k context window using DCA, achieving a PPL score of 5.59 \\
    \multicolumn{1}{r}{\model 13B} & The Llama2 70B model can be scaled up to 192k context with DCA. \\
    \multicolumn{1}{r}{Dynamic NTK 13B} & \cellcolor{gray!15} \verb|{\end{enumerate}\end{figure}[ht!]| 
    
    \verb|\end{document}\end{document}\begin{table}|\color{red}{\xmark (error)}
\\\cmidrule{1-2}
    \multicolumn{1}{r}{\textit{Question}} & In addition to Llama2, have the authors validated their method on other models? \\
    \multicolumn{1}{r}{\model {70B}} &  \cellcolor{gray!15} Yes, the authors have validated their method on other models, including 
    
    Together-32k, 
    
    CodeLlama, 
    
    which are all based on Llama. They have also compared their method to a variety of open-source and proprietary models.\\
    \multicolumn{1}{r}{\model 13B} & The authors have validated their method on the following models:
    
    * Together-32k
    
    * CodeLlama
    
    * Llama2 Long  \color{red}{\xmark (We did not verify DCA on Llama2 Long)}\\
    \multicolumn{1}{r}{Dynamic NTK 13B} & \cellcolor{gray!15} The paper introduces a new training-free framework called Dual Chunk Attention (DCA) that extends the context window by more than 8 times for 7B/13B Llama2 variants and 20 times for 70B models\color{red}{\xmark (ignore the instruction)}
 \\\cmidrule{1-2}
    \multicolumn{1}{r}{\textit{Question}} & Describe the finetuning corpus used by DCA\\
    \multicolumn{1}{r}{\model {70B}} &  \cellcolor{gray!15} The finetuning corpus used by DCA is a combination of long dialogue data from ShareGPT and AlpacaGPT4 \\
    \multicolumn{1}{r}{\model 13B} &  The finetuning corpus used by DCA is the same as the one used by the original Llama2 model, which is a combination of the training set and the validation set of the original Llama2 model   \color{red}{\xmark (We did not use training set of Llama)}\\
    \multicolumn{1}{r}{Dynamic NTK 13B} & \cellcolor{gray!15} \verb|{\end{begin}\begin{table}\end{figure}|
    
    \verb|[ht!]} \end{document}\end{document}\end{document}| \color{red}{\xmark (error)}\\
 \\\cmidrule{1-2}
    \multicolumn{1}{r}{\textit{Question}} & How many training steps does Llama2 Long need?\\
    \multicolumn{1}{r}{\model {70B}} &  \cellcolor{gray!15} Llama2 Long needs 100,000 training steps \\
    \multicolumn{1}{r}{\model 13B} &  Llama2 Long needs 100,000 training steps. \\
    \multicolumn{1}{r}{Dynamic NTK 13B} & \cellcolor{gray!15} Llama2 Long needs 40,000 steps for a 4k context window. \color{red}{\xmark (not 40,000)}\\
\\\cmidrule{1-2}
    \multicolumn{1}{r}{\textit{Question}} & Do the authors validate the model on L-Eval? and list the tasks\\
    \multicolumn{1}{r}{\model {70B}} &  \cellcolor{gray!15} Yes, the authors validate their model on L-Eval and list the tasks in Table 4. They evaluate their model on four closed-ended tasks from L-Eval with diverse input lengths ranging from 3k to 27k. The tasks include TOFEL, QuALITY, Coursera, and SFiction.\\
    \multicolumn{1}{r}{\model 13B} &   Yes, the authors do validate the model on L-Eval and list the tasks. In the section titled "Experiments", the authors present an extensive evaluation of their model on a diverse set of tasks, including NarrativeQA, Qasper, Quality, and QMSum.  \color{red}{\xmark (These are not from L-Eval)} \\
    \multicolumn{1}{r}{Dynamic NTK 13B} & \cellcolor{gray!15} The paper presents a new training-free framework called Dual Chunk Attention (DCA) that extends the context window by more than 8 times for 7B/13B Llama2 variants and... \color{red}{\xmark (ignore the instruction)}\\
\bottomrule
\end{tabular}
\end{table*}

\begin{table*}[t!]
    \small
    \centering
    \caption{QA (hard) results of \model 13B/70B and Dynamic-NTK with Llama2 13B, using the content of this paper as input. Questions in this table typically require readers to understand this method. The input prompt has 19388 tokens.  Dynamic-NTK is unable to give the correct answer for all test cases. Low-quality generation results are marked with \color{red}{\xmark}.}
    \label{tab:case2}
    \begin{tabular}{@{}c  p{0.8\textwidth}}
     \toprule
     \multicolumn{1}{r}{\textit{Prompt}} & [INST] $<<$SYS$>>$ Please read the long paper and  answer the questions $<</$SYS$>>$ 
   1. Introduction \textbackslash n\textbackslash n The ability to comprehend and process long-context information is essential for large language models (LLMs) to cater to a wide
   range of applications effectively. These include analyzing and responding to inquiries within sizable PDFs, retaining extended dialogue history, and empowering interactive chatbots... [/INST]\\
    \\\cmidrule{1-2}
    \multicolumn{1}{r}{\textit{Question}} & Why do the authors introduce successive-chunk attention? (Hard) \\
    \multicolumn{1}{r}{\model {70B}} &  \cellcolor{gray!15} The authors introduce successive-chunk attention to address the issue of \textbf{losing locality} in the attention mechanism when using inter-chunk attention. Inter-chunk attention can lead to a situation where the relative position between two tokens in different chunks is not accurately captured, resulting in a high perplexity. Successive-chunk attention helps to maintain locality by adjusting the position indices of the first $w$ tokens in each chunk to ensure that the relative position between neighboring tokens is accurately captured. This allows the model to better handle long-context tasks that require the ability to capture long-range dependencies.\\
    \multicolumn{1}{r}{\model 13B} &  The authors introduce successive-chunk attention to address the issue of long-range dependencies in long-context language models. The original positional encoding methods, such as RoPE and NTK-aware RoPE, do not capture long-range dependencies effectively, leading to poor performance in language modeling and passkey retrieval tasks. By introducing successive-chunk attention, the model can capture both local and long-range dependencies.. \color{red}{\xmark (misunderstand  successive-chunk attention)}  \\
    \multicolumn{1}{r}{Dynamic NTK 13B} & \cellcolor{gray!15} The paper introduces a new training-free framework called Dual Chunk Attention (DCA) that extends the context window by more than 8 times for 7B/13B Llama2 variants...\color{red}{\xmark (ignore the instruction)}\\
 
 \\\cmidrule{1-2}
    \multicolumn{1}{r}{\textit{Question}} &  Why do the authors introduce inter-chunk attention? (Hard) \\
    \multicolumn{1}{r}{\model {70B}} &  \cellcolor{gray!15} The authors introduce inter-chunk attention to address the issue of poor \textbf{long-range dependencies} in the intra-chunk attention module, which uses a fixed context window for attention computation. The inter-chunk attention mechanism allows the model to attend to tokens from different chunks, enabling it to capture long-range dependencies and improve performance on tasks that require processing long sequences.\\
    \multicolumn{1}{r}{\model 13B} &  13b: The author introduces inter-chunk attention to improve the passkey retrieval performance of the model. The author notes that using only intra-chunk attention, which disregards information from previous chunks, can maintain a low PPL but hinder the model's ability to retrieve passkeys from other chunks. \color{red}{\xmark  (not accurate enough)}  \\
    \multicolumn{1}{r}{Dynamic NTK 13B} & \cellcolor{gray!15} The paper presents a new training-free framework called Dual Chunk Attention (DCA) that extends the context window by more than 8 times for 7B/13B Llama2 variants...\color{red}{\xmark (ignore the instruction)}\ \\
 \\\cmidrule{1-2}
    \multicolumn{1}{r}{\textit{Question}} & Why the relative position in DCA will not exceed the pretraining context length? (Hard)\\
    \multicolumn{1}{r}{\model {70B}} &  \cellcolor{gray!15} The relative position in DCA is calculated using the position indices of the keys and queries, which are both less than or equal to the pretraining context length. Therefore, the relative position will not exceed the pretraining context length.  \color{red}{(not clear enough)} \\
    \multicolumn{1}{r}{\model 13B} &  The relative position in DCA will not exceed the pretraining context length because the model only considers the relative positions within the same chunk. The chunks are segmented based on the pretraining context length, and the relative positions are calculated within each chunk.   \color{red}{(not clear enough)}\\
    \multicolumn{1}{r}{Dynamic NTK 13B} & \cellcolor{gray!15} The relative position in Dual Chunk Attention (DCA) does not exceed the pretraining context length because it is designed to maintain locality. \color{red}{(not clear enough)}\\
 
 \\\cmidrule{1-2}
    \multicolumn{1}{r}{\textit{Question}} & What are the key differences between this paper and earlier papers? (Hard)\\
    \multicolumn{1}{r}{\model {70B}} &   The key differences between this paper and earlier papers on large language models (LLMs) are:

1. The paper proposes a new method called Dual Chunk Attention (DCA) that enables LLMs to process long-context information without additional training costs.

2. DCA is designed to handle long-context information by segmenting the input sequence into smaller chunks and computing attention within each chunk, which reduces the computational cost and allows for efficient scaling of the context window... \\
    \multicolumn{1}{r}{\model 13B} & \cellcolor{gray!15}  This paper proposes a new training-free framework called Dual Chunk Attention (DCA) that enables large language models (LLMs) to process long-context information without additional training. DCA introduces a new attention mechanism that scales the context window of LLMs without additional training, allowing for more efficient use of computational resources.  \\
    \multicolumn{1}{r}{Dynamic NTK 13B} &  DCA is orthogonal to popular extrapolation methods such as PI and NTK-Aware RoPE, and can be integrated with existing long-context models to scale up significantly...  \\

\bottomrule
\end{tabular}
\end{table*}

\end{document}